\documentclass[runningheads]{llncs}
\usepackage[T1]{fontenc}
\usepackage{graphicx}

\usepackage{amsmath}
\usepackage{amssymb}
\usepackage{booktabs}
\usepackage{multirow}
\usepackage{enumitem}
\usepackage{array} 
\usepackage{tabularx}
\usepackage{fontawesome}
\usepackage{hyperref}
\usepackage{color}
\usepackage{bm}

\usepackage{array,booktabs}
\newcommand {\otoprule}{\midrule [\heavyrulewidth]}
\newcolumntype {+}{ >{\global\let\currentrowstyle\relax}}
\newcolumntype {^}{ >{\currentrowstyle }}
 \newcommand {\rowstyle}[1]{\gdef\currentrowstyle{#1} %
 #1\ignorespaces
 }
\newcommand{\tabhead}{\rowstyle{\bfseries}}

\usepackage[table]{xcolor} 
\definecolor{coluser}{RGB}{51, 187, 238}
\definecolor{colpresent}{RGB}{238, 119, 51} 
\definecolor{colcontent}{RGB}{0, 153, 136}
\makeatletter
\newcolumntype{i}[1]{%
    >{\minipage[t]{\linewidth}\let\\\tabularnewline
      \itemize
      \addtolength{\rightskip}{0pt plus 50pt}
      \setlength{\itemsep}{-\parsep}}%
    p{#1}%
    <{\@finalstrut\@arstrutbox\enditemize\endminipage}}
\makeatother

\begin{document}
\title{The Co-12 Recipe for Evaluating Interpretable Part-Prototype Image Classifiers}
\titlerunning{Evaluating Interpretable Part-Prototype Image Classifiers}
%
\author{Meike Nauta\inst{1}\orcidID{0000-0002-0558-3810} \and
Christin Seifert\inst{2,3}\orcidID{0000-0002-6776-3868}}
\authorrunning{M. Nauta et al.}
%
\institute{University of Twente, Enschede, the Netherlands\\
\email{m.nauta@utwente.nl}\\
\and
University of Duisburg-Essen, Germany\\
\and
University of Marburg, Germany\\
\email{christin.seifert@uni-marburg.de}}
\maketitle              
\begin{abstract}
Interpretable part-prototype models are computer vision models that are explainable by design. The models learn prototypical parts and recognise these components in an image, thereby combining classification and explanation. Despite the recent attention for intrinsically interpretable models, there is no comprehensive overview on evaluating the explanation quality of interpretable part-prototype models. Based on the Co-12 properties for explanation quality  as introduced in~\cite{Nauta2023_csur_evaluating-xai-survey} (e.g., correctness, completeness, compactness), we review existing work that evaluates part-prototype models, reveal research gaps and outline future approaches for evaluation of the explanation quality of part-prototype models. This paper, therefore, contributes to the progression and maturity of this relatively new research field on interpretable part-prototype models. We additionally provide a ``Co-12 cheat sheet'' that acts as a concise summary of our findings on evaluating part-prototype models.

\end{abstract}

\section{Introduction}
The goal of Explainable AI (XAI) is to make the reasoning of a machine learning model accessible to humans, such that users of an AI system can understand the underlying model~\cite{barredo_arrieta_explainable_2020}. Over the last years, many methods and approaches to explain (mostly deep) learning models were proposed~\cite{guidotti2018survey}.

A machine learning model maps an input $x$ to an output $\hat{y}$, and can be described as function $\hat{y}=f(x)$. XAI develops explanation methods $e$ for machine learning models, thus an XAI method represents $e(f(x))$. In the case of intrinsically interpretable models, such as decision trees, $e$ equals the predictive model $f$ since $e(x)=f(x)$. The key aspect of this formalism is that, generally, the user facing output is a combination $e(f)$ of model and explanation. While the community has built a de-facto standard for evaluating machine learning models (e.g., cross-validation, train/validation/test splits, standard evaluation metrics), there is no common agreement on the evaluation of XAI methods~\cite{Nauta2023_csur_evaluating-xai-survey}.

\begin{figure}[!t]
    \centering
    \includegraphics[width=0.99\textwidth]{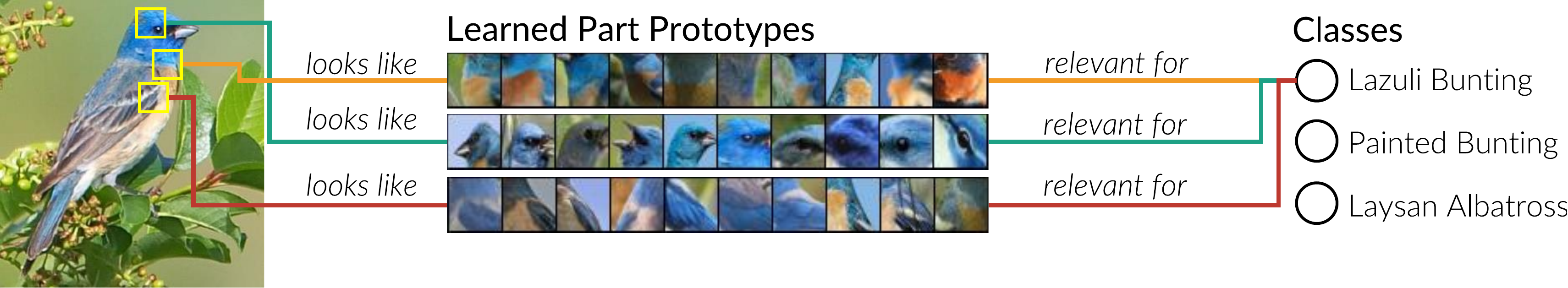}
    \caption{Part-prototype models learn prototypical parts that can be visualised as image patches. The decision layer learns which prototypes are relevant for which classes. Classification depends on the presence of part-prototypes in an input image.}
    \label{fig:ppmodel}
\end{figure}

A promising type of interpretable-by-design models are part-prototype models, which base their reasoning on the recognition-by-components theory~\cite{biederman1987recognition}. Specifically, part-prototype models use deep neural networks to learn semantically meaningful components (\emph{prototypical parts}), without relying on additional part annotations. Subsequently, an image is classified by automatically testing whether patches in an input image look similar to a learned prototypical part. Thus, the presence or absence of specific parts in the image determines the decision process, as visualised in Figure~\ref{fig:ppmodel}. Part-prototype models are globally interpretable since the learned part-prototypes can be visualised as image patches, and their decision layer is an interpretable model linking prototypes to classes, such as a decision tree~\cite{Nauta_2021_CVPR_ProtoTree} or linear layer~\cite{nips/ChenLTBRS19,nauta_pipnet}. Apart from their development using benchmark data, part-prototype models are applied in the medical domain, e.g., on X-rays~\cite{Singh2021-proto-on-chest-xray}, MRI-scans~\cite{mohammadjafari2021using}, mammograms~\cite{carloni_2022,wang2022knowledge} or CT-scans~\cite{Singh2022-proto-on-chest_ctscan}.

The majority of XAI evaluation methods  are designed for explanation types produced by post-hoc XAI methods, which separate $e$ from $f$, such as heatmaps and feature importance scores~\cite{Nauta2023_csur_evaluating-xai-survey}. In contrast, there is little work on evaluating interpretable-by-design models where $e=f$, as also observed by~\cite{Kim2022HIVE}.

\begin{table}[!t]
  \caption{Summary of the Co-12 explanation quality properties~\cite{Nauta2023_csur_evaluating-xai-survey}.}
  \label{tab:desiderata_overview}
  \centering
  \begin{tabular}{p{0.2\linewidth}p{0.75\linewidth}}
  \toprule\tabhead
    Property & \textbf{Describes ...} \\\otoprule
    \rowcolor{colcontent}
    \multicolumn{2}{c}{\sc Content}\\
    Correctness & how faithful the explanation is w.r.t.~prediction model\\
    {Completeness} & how much of the model behaviour is described in the explanation\\
    {Consistency} & how deterministic and implementation-invariant the explanation method is\\
    {Continuity} & how continuous and generalisable the explanation is\\   
    {Contrastivity} & how discriminative the explanation is w.r.t.~other events\\
    {Covariate\newline complexity} & how complex the (interactions between) features in the explanation are\\
    \midrule
    \rowcolor{colpresent} 
    \multicolumn{2}{c}{\sc Presentation}\\
    {Compactness} & the size of the explanation \\
    {Composition} & the presentation and organisation of the explanation \\
    {Confidence} & the presence and accuracy of probability information in the explanation\\
        \midrule
    \rowcolor{coluser}
   \multicolumn{2}{c}{\sc User}\\
    {Context} & how relevant the explanation is to the user\\
    {Coherence} & how accordant the explanation is with prior knowledge and beliefs\\
    {Controllability} & how interactive or controllable the explanation is\\
     \bottomrule
  \end{tabular}
\end{table}

In this paper, we collected work that evaluates interpretable part-prototype models and summarise its findings. Moreover, we outline future approaches for evaluating the explanation quality of part-prototype models. For a comprehensive view on the quality of an explanation (method), we organise the existing and suggested evaluation methods for part-prototype models per Co-12 property as introduced in~\cite{Nauta2023_csur_evaluating-xai-survey} and outlined in Table~\ref{tab:desiderata_overview}. The Co-12 properties are a high-level decomposition of explanation quality, including  Correctness and Compactness. By addressing the Co-12 properties individually, we identify research gaps and we conceptualise how the Co-12 properties can be put in practice for a thorough evaluation of interpretable image classifiers based on prototypical parts, including ProtoPNet~\cite{nips/ChenLTBRS19}, ProtoTree~\cite{Nauta_2021_CVPR_ProtoTree}, ProtoPShare~\cite{protopshare_rymarczyk_2021}, ProtoPool~\cite{rymarczyk_2022_protopool}, TesNet~\cite{Wang_2021_ICCV_tesnet}, ProtoPFormer~\cite{xue2022protopformer} and PIP-Net~\cite{nauta_pipnet}. We provide an overview of existing methods to evaluate part-prototype models and hope that our identification of future research opportunities serves as a `recipe' for further evaluation. 

\paragraph{Paper Outline:} In the following, we review each Co-12 evaluation property in the order of Table~\ref{tab:desiderata_overview}. We dedicate one section to each property, starting with a general description, reviewing related work w.r.t. this property and finally derive evaluation recommendations. We discuss the evaluation of part-prototype models from a broader perspective in Section~\ref{sec:discussion} and conclude our work in Section~\ref{sec:conclusion}.

\section{Evaluating Co-12 Properties}
\label{sec:prop}
In this section, we discuss the Co-12 properties, review related work for each property and provide recommendations towards a more comprehensive evaluation of interpretable part-prototype models.
Table~\ref{tab:co12_cheat_sheet_proto} supports readers by providing a ``Co-12 cheat sheet'' that acts as a concise summary of our findings on evaluating part-prototype models.
\begin{table*}
\centering
\caption{``Cheat sheet'' for evaluating part-prototype models along Co-12 properties.}
\label{tab:co12_cheat_sheet_proto} 
\rowcolors{2}{gray!7}{white}
\begin{tabular}{>{\centering}p{1.4cm}p{2cm}p{0.7\linewidth}}
\toprule
     \multicolumn{2}{l}{\textbf{Property}}  & \textbf{Evaluation Approaches} \\\otoprule
     \rowcolor{colcontent}
    \multicolumn{3}{c}{\sc Content}\\
     {\color{colcontent}\small\textbf{$\bm{f\stackrel{?}{=}e}$}}
        & Correctness 
        & \multicolumn{1}{i{0.7\linewidth}}{
        \item Classification process of part-prototype models is correct by design since $f(x)=e(x)$
        \item Evaluate prototype visualisation with synthetic data or incremental deletion/addition of image patches
        } \\
    {\color{colcontent}\large\faPuzzlePiece}
        & Completeness
        & \multicolumn{1}{i{0.7\linewidth}}{
        \item Output-complete by design
        \item Evaluate human-output-completeness with simulatibility user studies
        } \\
    {\color{colcontent}\small$\bm{e=e}$}
        & Consistency
        & \multicolumn{1}{i{0.7\linewidth}}{
        \item Implementation invariance and nondeterminism
        }\\
    {\color{colcontent}\small$\bm{e \thickapprox e}$}    
        & Continuity
        & \multicolumn{1}{i{0.7\linewidth}}{
        \item Stability for slight variations 
        }\\
    {\color{colcontent}\small$\bm{e}$ $\bm{\leftrightarrow}$ $\bm{e'}$}
        & Contrastivity
        &  \multicolumn{1}{i{0.7\linewidth}}{
        \item Contrastive by design; can answer counterfactual questions
        \item Pragmatism and compactness for optimal contrastive explanation
        \item Target-sensitivity for location of prototypes
        \item Target-discriminativeness to evaluate prototypes
        } \\
     {\color{colcontent}\footnotesize$\square\blacklozenge\triangle$   \footnotesize$\square\blacksquare\square$}
        & Covariate complexity 
        &\multicolumn{1}{i{0.7\linewidth}}{
       \item Prototype homogeneity / purity with annotated data
       \item Perceived homogeneity with user studies (subjective)
       \item Intruder detection for objective homogeneity evaluation
        } \\
    \rowcolor{colpresent} 
    \multicolumn{3}{c}{\sc Presentation}\\
    {\color{colpresent}\large\faCompress} 
        & Compactness
        & \multicolumn{1}{i{0.7\linewidth}}{
        \item Number of prototypes (local \& global)
        \item Number of unique prototypes  (redundancy)
        }\\
    {\color{colpresent}\faPaintBrush}
        & Composition
        & \multicolumn{1}{i{0.7\linewidth}}{
        \item Compare different explanation formats with the same content
        \item User study on how to present part-prototypes
        \item User study on classification format (e.g. linear layer or decision tree)
        }\\
     {\color{colpresent}\large\faBarChartO}
        & Confidence
        & \multicolumn{1}{i{0.7\linewidth}}{
        \item Reliability of classification confidence
        \item Reliability of explanation confidence
        \item Out-of-distribution detection confidence
        }\\
    \rowcolor{coluser}
   \multicolumn{3}{c}{\sc User}\\
    {\color{coluser}\large\faUserMd}
        & Context
        & \multicolumn{1}{i{0.7\linewidth}}{
        \item User studies (lab and field)
        } \\
    {\color{coluser}\large\faThumbsUp}
        & Coherence
        & \multicolumn{1}{i{0.7\linewidth}}{
        \item Anecdotal evidence by visualising reasoning with prototypes
        \item Alignment with domain knowledge from annotated data
        \item Subjective satisfaction with user studies
        } \\
    {\color{coluser}\large\faWrench} 
        & Controllability
        & \multicolumn{1}{i{0.7\linewidth}}{
        \item GUI for interactive and personalised explanations
        \item Explanatory debugging to manipulate prototypes and model's reasoning
        } \\
        \bottomrule
\end{tabular}
\end{table*}

\subsection{Correctness}
\label{sec:prop:correctness}
Part-prototype models are interpretable by design: explanation method $e$ is incorporated in predictive model $f$. Since a part-prototype model is simultaneously making predictions \emph{and} providing explanations, the correctness of the reasoning process 
is fulfilled by design. The only part of the explanation that is not guaranteed to be faithful to $f$ is the prototype visualisation for converting latent representations to visual natural image patches. To evaluate the correctness of the visualisation in part-prototype models, one could apply a \emph{controlled synthetic data check}~\cite{Nauta2023_csur_evaluating-xai-survey,liu2021synthetic} where the discriminative features are known a-priori. If one can safely assume that the model is following a particular reasoning (e.g. when the classifier has near-perfect accuracy), it can be checked whether the prototypes reveal the ground-truth important features. Additionally, correctness can be evaluated with \emph{single deletion} where one important image patch or set of pixels is removed, or \emph{incremental deletion} or \emph{addition}, where image pixels or patches are removed one by one. Correct prototype importance scores should result in an Area under the Deletion Curve (AUDC) that is lower than an AUDC for random rankings. With incremental \emph{addition}, image pixels or patches are added one by one to an initially empty image based on their similarity with a prototype. When starting with the image patch with the highest similarity to a prototype, only one image patch could already be sufficient to activate that prototype and therefore influence the reasoning of the model.

\subsubsection{Related Work.} Yeh et al.~\cite{nips/YehKALPR20} applied the controlled synthetic data check by constructing a set of synthetic images with coloured shapes, and only a few of those shapes were relevant for the ground-truth class. Since their interpretable concept-based classifier obtained near-perfect accuracy, it could be safely assumed that the model had learned the intended reasoning, such that it could be evaluated whether those discriminative shapes are indeed included in the explanation. Similarly, Gautam et al.~\cite{gautam_prp_2023} added a yellow square to images to evaluate whether prototypes reveal the added artefact. 

Xu-Darme et al.~\cite{xudarme_romain_sanity_2023} applied the incremental deletion metric to ProtoTree~\cite{Nauta_2021_CVPR_ProtoTree} and ProtoPNet~\cite{nips/ChenLTBRS19}. They find that the AUDC is decreasing when more pixels are removed, and that the decline is steeper for the improved visualisation method Prototypical Relevance Propagation (PRP)~\cite{gautam_prp_2023}, hence confirming the faithfulness of part-prototype explanations. Compared to ProtoPNet, the AUDC drops slower for ProtoTree, and the authors  hypothesise that this could be explained by the fact that ProtoTree shares prototypes between classes and therefore focuses less on small details~\cite{xudarme_romain_sanity_2023}. 
Gautam et al.~\cite{gautam_prp_2023} perform the incremental addition test for ProtoPNet, starting from a random image and incrementally adding the most relevant pixels for a prototype. They find that similarity scores for a prototype indeed increase after adding more pixels, and that their improved visualisation approach, called PRP~\cite{gautam_prp_2023}, results in a steeper slope, implying that fewer pixels are needed to obtain a certain similarity score and that the visualisation with PRP is more faithful to the model's reasoning. 

\subsubsection{Recommendations.} Although part-prototype models are interpretable by design, the correctness of approaches for \emph{visualising} part-prototypes can be further evaluated. In addition to the original visualisation approach with bicubic upsampling, we recommend to evaluate new methods, such as the recently introduced PRP method~\cite{gautam_prp_2023} for more precise localisation of prototypical parts. Controlled Synthetic Data Checks could be performed to evaluate the correctness of the visualised prototypes. In addition to simplistic synthetic data, more advanced datasets can be used for this purpose, such as Animals with Attribute~\cite{lampert_animals_attributes}, BAM~\cite{BAM2019}, NICO~\cite{he2021towards_nico} and a recent set of synthetic benchmarks for XAI~\cite{liu2021synthetic}.  Evaluating with synthetic data checks may in turn lead to further insights for improving prototype visualisation. 

We also recommend to apply existing deletion and addition methodology, usually applied to heatmaps, to part-prototype models. Important here is to take the specific nature of part-prototype models into account. Features in prototype explanations are not single pixels, but rather object parts that can occur more than once in an image. For instance, a prototype that represents a car tyre will still be detected when only one of four tyres is masked. Therefore, not only the image pixel/patch with the highest similarity score to the prototype should be removed or perturbed, but \emph{all} pixels/patches with a sufficiently high similarity score to that particular prototype. Additionally, an often-raised criticism for deletion methods is that naively removing or perturbing image pixels in an input image could lead to out-of-distribution (OoD) data. Moreover, it was found that the \emph{shape} of a mask could leak class information to the model~\cite{pmlr-v162-rong22a,xie2022vit}. Various approaches are recently presented to circumvent these OoD and shape-leakage issues, e.g.~\cite{pmlr-v162-rong22a,hase_nips_21_ood_xai}, and we encourage the research community to adapt and apply these improved evaluation methods to part-prototype models. 

\subsection{Completeness}
\label{sec:prop:completeness}
Completeness describes how much of the black box behaviour is described in the explanation.
We can distinguish between \emph{reasoning}-completeness, which is generally not quantified but rather a property of the model, and \emph{output}-completeness. Reasoning-completeness indicates the extent to which the explanation describes the entire \emph{internal} workings of the model. Part-prototype models transparently show the relation between prototypes and classes, but intentionally abstract away all matrix multiplications and other calculations of the neural network backbone by showing prototypical parts. Output-completeness addresses the extent to which the explanation covers the output of a predictive model $f$. Evaluating output-completeness is highly relevant for post-hoc explanation methods where explanation method $e$ is applied to a trained model $f$, as it measures whether the explanation is sufficient for explaining the output of model $f$. Instead, prototypical part models are designed to be intrinsically interpretable, such that $e=f$, implying that output-completeness is fulfilled by design. 

One could also evaluate \emph{human-}output-completeness with the \emph{forward simulatability} evaluation method~\cite{Nauta2023_csur_evaluating-xai-survey}, which measures whether a user can simulate the model's reasoning by letting the user predict the model's output based on the input and the explanation. A human-output-complete explanation with understandable prototypes should be sufficient for a user to follow the model's reasoning and hence end up with the same prediction. Also the \emph{counterfactual simulatability} can be measured~\cite{doshi-velez_considerations_2018}, where the user is given the input, explanation and output, after which the user should predict what the model will output for a perturbed version of the initial input. 

\subsubsection{Related Work.}
Hase \& Bansal~\cite{acl/HaseB20} evaluated the forward and counterfactual simulatability of a prototype model for text and tabular data, and found that prototype models are effective in counterfactual simulation tests. Although the prototypes were full data samples rather than prototypical \emph{parts}, and not evaluated on image data, their result is promising and a motivation for further studies on human-output-completeness of part-prototype models. 
Kim et al.~\cite{Kim2022HIVE} evaluated the forward simulatability of part-prototypes~(ProtoTree and ProtoPNet) and heatmaps (BagNet~\cite{bagnet_iclr} and GradCAM~\cite{iccv/SelvarajuCDVPB17}). They conclude that participants have difficulties in predicting the model's output based on the explanations, as participants predict between 43\% (for ProtoTree) to 48\% (for GradCAM) of the outputs correctly when the model predicted the correct class label, and 33\% (BagNet and ProtoTree) to 36\% (GradCAM) when the model was incorrect. 

\subsubsection{Recommendations.} Model-output-completeness is fulfilled by design in part-prototype models.
Simulatability studies for measuring human-output-complete-ness is related to other Co-12 properties such as Correctness, Covariate complexity and Coherence. We think that the forward simulatability performance for part-prototype models can be improved by addressing other Co-12 properties. Specifically, 
improving the Correctness of the visualisation (e.g. by applying recently introduced visualisation methods that locate prototypes more precisely~\cite{gautam_prp_2023,xudarme_romain_sanity_2023}) and the Covariate complexity of the visualised prototypes, such as showing multiple image patches rather than one~(c.f.~Sect.~\ref{sec:prop:cov-complexity}) can have a positive impact on human-output-completeness.

\subsection{Consistency}
\label{sec:prop:consistency}
Consistency addresses the idea that identical inputs should have identical explanations and that explanations should be implementation invariant. 
Where most XAI publications already report predictive performance across multiple runs, this property focuses on the consistency of \emph{explanations}. 
Existing part-prototype models are in principle deterministic as there is no source of randomness built into their design. At test time, identical images should thus result in identical explanations. However, even though part-prototype models do not have intentionally added random components in their designs, nondeterminism can still occur from their neural network backbones, such as model initialisation and random seeds. 
The evaluation method \emph{implementation invariance} quantifies consistency in more detail, in order to analyse experiment repeatability~\cite{Nauta2023_csur_evaluating-xai-survey,montavon_gradient-based_2019}.

\subsubsection{Related Work.}
The implementation invariance of explanations from part-prototype models can be evaluated by comparing explanations from models trained with different initialisations or with a different shuffling of the training data. For example, the latent distances of prototypes in ProtoTree~\cite{Nauta_2021_CVPR_ProtoTree} are evaluated across different runs. \cite{rymarczyk_2022_protopool} specifically analysed for ProtoPool the distribution of how prototypes are shared between classes and reported negligible differences between different runs. 

\subsubsection{Recommendations.} Implementation invariance is not only important for explanation consistency, but also for experiment replicability in order to fairly validate part-prototype models and to be able to quantify improvements of new methods. While the most salient factors of nondeterminism are model initialisation and random seeds, it is important to be aware of other sources of randomness, including data augmentation, shuffling of training data, dropout, batch normalisation, tooling and hardware~\cite{summers_nondeterminism_21,zhuang_randomness_22}. \cite{zhuang_randomness_22} report for standard neural networks that top-line evaluation metrics as top1-accuracy are not much impacted by randomness, but that models are far more sensitive on certain parts of the data distribution. For future work, we think it would be interesting to quantify to what extent classification decisions and explanations are influenced by nondeterminism in part-prototype models, taking inspiration from existing experiments on nondeterminism and randomness~\cite{summers_nondeterminism_21,zhuang_randomness_22}. Additionally, a comparison can be made between the implementation invariance of standard black box classifiers and interpretable part-prototype models. Lastly, we would like to emphasise that most of the nondeterminism can be eliminated with the right code implementations, although with potentially significant overhead~\cite{zhuang_randomness_22}. Ensuring Consistency will therefore be a trade-off with computation speed and resources, and may depend on the criticality of the application. 

\subsection{Continuity}
\label{sec:prop:continuity}
Continuity describes how continuous and generalisable the explanation function is, targeting that similar inputs should have similar explanations.
For part-prototype models, \emph{stability for slight variations}~\cite{Nauta2023_csur_evaluating-xai-survey} is relevant, as it quantifies whether slightly perturbed inputs lead to the same explanation, given that the model makes the same classification. Such stability is especially relevant for generalisability, as input data might be collected from different data sources with slightly different image characteristics~\cite{hoffmann2021looks}. 

\subsubsection{Related Work.} 
The evaluation method \emph{stability for slight variations} was implemented by~\cite{hoffmann2021looks}  with image compression and adding adversarial human-impercep-tible noise to images, after which they found that ProtoPNet~\cite{nips/ChenLTBRS19} assigns drastically lower similarity scores to prototypes for these perturbed images. They conclude that there is a semantic gap between ProtoPNet's understanding of similarity (in latent space) and that of humans (in input space). To address this shortcoming, Hoffmann et al.~\cite{hoffmann2021looks} found that the continuity of ProtoPNet can be improved with adversarial training and data augmentation, albeit at the cost of classification accuracy. Adding a regularisation mechanism to ProtoPNet can also improve its robustness to adversarial attacks~\cite{nakka_robust_2020}. 
The ``transformation invariance'' of ProtoTree and ProtoPNet is evaluated in~\cite{sinhamahapatra2022towards}, arguing that prototypes should capture a semantic meaningful concept ``irrespective of their variability in scale, translation, or rotation angle across different samples''. 
They evaluated transformation invariance by forwarding a cropped or rotated test image through the prototype model and analysing whether similar prototypes are found. They found that ProtoPNet detects different prototypes, even from different classes, in the transformed test sample, whereas the prototypes and hence the decision path in ProtoTree did not change after the transformation, ``indicating ProtoTree to be more robust to image transformations than ProtoPNet''~\cite{sinhamahapatra2022towards}. Rymarczyk et al.~\cite{protopshare_rymarczyk_2021} evaluated ProtoPNet's and their ProtoPShare model's resistance to perturbations by modifying brightness, contrast, saturation, hue and perspective of images. They found that the accuracy of both models decreases only slightly for large perturbations. 

PIP-Net incorporates continuity already \emph{during} training with its contrastive learning of prototypes. PIP-Net is optimised to assign the same prototype similarity scores to two augmented views of the same image. As such, human perception of similarity is indirectly encoded in the data augmentation, and thus in the training process. Also noise and image compression as used by \cite{hoffmann2021looks} can be included in PIP-Net's data augmentation. 

\subsubsection{Recommendations.} 
Continuity of predictive models is important to ensure their robustness and generalisability. It also contributes to the predictability of explanations and may in turn influence the user's trust in part-prototype models. Recent work has shown that the continuity of prototype-based explanations can be improved with extended data augmentation and adversarial training, and we therefore recommend to include image variations that might occur in the intended application in the training process of part-prototype models. Since adversarial attacks and defence mechanisms are an active research area~\cite{michel2022survey}, we also support further research that would investigate whether more advanced adversarial defence methods can be incorporated in the training process of part-prototype models. Follow-up research could then analyse whether prototype-based models are more continuous and robust than standard black box classifiers. 

\subsection{Contrastivity}
\label{sec:prop:contrastivity}
Contrastivity describes how discriminative the explanation is with respect to other events or targets, and contrastive explanations help to answer counterfactual questions as ``why not?'' and ``what if?''. 
The \emph{target sensitivity} evaluation method captures that explanations should differ between classes and is mostly evaluated for heatmaps~\cite{Nauta2023_csur_evaluating-xai-survey}, since a heatmap for one class should be different from a heatmap that explains another class~\cite{adebayo_sanity_2018,rudin2019stop}.
Part-prototype models have their interpretability incorporated into their prediction model, so the local explanation is by design output-complete~(c.f.~Section~\ref{sec:prop:completeness}). As a result, a different classification corresponds to a different reasoning and hence to a different explanation. Contrastivity is thus by design incorporated in part-prototype models. 

\subsubsection{Related Work.} 
Sinhamahapatra et al.~\cite{sinhamahapatra2022towards} evaluated \emph{subjective} target discriminativeness with a small-scale user study by analysing whether users could guess the class that a set of prototypes belongs to. 
The class prediction accuracy for ProtoPNet was higher than ProtoTree (98\% vs 55\%), which is, as the authors also note, expected since ProtoPNet learns class-specific prototypes, whereas ProtoTree's prototypes are shared between classes such that the set of prototypes on one decision path will not necessarily correspond to a single class~\cite{sinhamahapatra2022towards}. Goyal et al.~\cite{icml/GoyalWEBPL19} specifically developed an approach for answering counterfactual questions by generating images in which a specific region is adapted such that the model's classification is changed. Whereas they do an exhaustive search to find the best counterfactual spatial region, we think that the prototypical parts learned by part-prototype models inherently provide the model's discriminative regions. They report that presenting counterfactual visualisations in \emph{addition} to representative training samples helps teaching users in classification tasks~\cite{icml/GoyalWEBPL19}. 

\subsubsection{Recommendations.} 
Contrastivity is inherently incorporated in part-prototype models. But rather than looking at \emph{which} prototypes are found in an input image, we see a research opportunity to also evaluate target-sensitivity by analysing \emph{where} prototypes are detected. For perfect target sensitivity, a part-prototype model should find different prototypes at different locations in the test image. E.g., a prototype of a bird's red beak should be detected at a different location than a prototype of a long tail. A fruitful idea for future work would be to analyse 
to what extent evidence for one prototype is found at a different location in a test image than evidence for another prototype, averaged over all images in a test set. 
Additionally, contrastivity can be measured with the \emph{target discriminativeness} method, which captures whether another, external model can predict the right class label from the explanation. It was found that this method is mostly applied to generative explanation methods that produce for example text or synthesised images~\cite{Nauta2023_csur_evaluating-xai-survey}. We see multiple ways to automatically quantify the target discriminativeness of prototypes visualised as image patches. The similarity scores per prototype can be organised in a tabular format, and provided as input to standard machine learning algorithms, such as linear classifiers. A high classification accuracy of the external model would indicate that the prototype similarity scores contain relevant discriminative information about the target. Alternatively, the prototype image patches can be provided to an external neural network, after which the deep classifier should be able to predict either the right prototype or the right class label. Interestingly, if the external classifier finds that a prototype is not discriminative for the target, it might be an indication that the prototype is also unnecessary in the part-prototype model. Removing it from the part-prototype model might then improve the compactness of the explanation, potentially without hurting predictive performance. 

Lastly, the built-in contrastivity also allows to answer counterfactual questions by identifying which prototypes should have been present or absent in an image for a different decision. E.g., a user can follow a decision path in ProtoTree bottom-up, starting at the actual class leaf, and find the crucial node that would have led to ending up in a different leaf and hence a different classification. Given the promising results of counterfactuals in a previous user study~\cite{icml/GoyalWEBPL19}, we encourage further experiments that use part-prototype models to generate counterfactual explanations. Since there are usually different counterfactual explanations available, the cost or size of the counterfactual change can be considered, as quantified by e.g. \emph{pragmatism} and \emph{counterfactual compactness}~\cite{Nauta2023_csur_evaluating-xai-survey}. 

\subsection{Covariate complexity}
\label{sec:prop:cov-complexity}
Covariate complexity describes how complex it is for users to understand the features in an explanation. 
In part-prototype models, the prototypical parts visualised as image patches are the explanation covariates. Quantifying the complexity or understandability of these visual features requires evaluation with human subjects, or relying on automated evaluation based on annotations made by humans. 
The main approach for evaluating the covariate complexity of prototypes is to evaluate their \emph{homogeneity}, which indicates how consistently a prototype represents a human-interpretable concept. In part-prototype models, a prototype ideally only gets high similarity scores for one semantically meaningful concept. Hence, the purity of the cluster with all image patches that have a high similarity with a  particular prototype should be high. Lakkaraju and Leskovec~\cite{nips/LakkarajuL16} also aim for what they call ``inverse purity'', meaning that ideally all image patches that encode a particular meaningful concept are in one cluster, indicating the cluster's completeness. These terms were introduced for evaluating clusters for time series data, where a prototype was defined as a representative data point for a cluster~\cite{nips/LakkarajuL16}, but we think that the purity and inverse purity metrics are also applicable to clusters with image patches. Evaluation will require a ground-truth, either provided by human judgements (\emph{perceived homogeneity}) or with annotations, such as predefined concepts~\cite{cvpr/FongV18} or object part annotations. 

Applying the \emph{intruder detection} evaluation method to prototype visualisations would be a more objective evaluation of perceived homogeneity. In this user study, participants are shown a set of image patches that all, except one, have a high similarity to one particular prototype. If a prototype is homogeneous and interpretable, participants should be able to discover the odd one out. 

\subsubsection{Related Work.} 
Borowski et al.~\cite{iclr/BorowskiZSGWBB21} found that exemplary natural images are more informative than generated synthetic images for predicting CNN activations, both for lay and expert participants~\cite{iclr/BorowskiZSGWBB21}. Additionally, participants were also faster and more confident for natural images. Their results are already a promising indication that image patches are understandable to humans. Sinhamahapatra et al.~\cite{sinhamahapatra2022towards} evaluate with manual inspection how well learned part-prototypes of ProtoTree and ProtoPNet correspond to a ``distinct human-relevant entity''. They confirm earlier findings~\cite{Nauta_2021_CVPR_ProtoTree} that prototypes in the top layers of the tree can be difficult to interpret as they are shared between many classes~\cite{sinhamahapatra2022towards}. For ProtoPNet, they report that understanding the meaning of a single image patch can be difficult, but that the context where prototypes are located improves the understanding. The method from~\cite{nauta_exproto_2021} addresses these ambiguities and can clarify the meaning of a prototype by quantifying the influence of colour hue, shape, texture, contrast and saturation. Such an additional explanation of what the prototype represents can thus reduce covariate complexity. 

To quantify covariate complexity, the purity of PIP-Net's prototypes can be measured by calculating whether the same ground-truth object part is shown in a prototype's top-10 most similar image patches, reporting a purity of up to 93\% for the CUB birds dataset~\cite{nauta_pipnet}. Ghorbani et al.~\cite{nips/GhorbaniWZK19} instead did an intruder detection experiment with human subjects for evaluating the coherency of learned visual concepts, and found that 97\% to 99\% of the provided answers were correct. They also asked participants to describe a set of image segments, representing a learned \emph{concept}, with one word and evaluated how many participants provided the same word. They found that 56\% of the participants described the concept with the most frequent word and its synonyms, and therefore concluded that the learned concepts are semantically and verbally meaningful to humans. Das et al.~\cite{icdm/DasXDER20} asked participants to select one out of six image patches that best explains an image and its class label. They then compared to users' selections with their ProtoPFormer's model selections and found that most of the post-hoc prototypes generated by their surrogate model are in line with human preferences, thereby concluding that their prototypes are human understandable~\cite{icdm/DasXDER20}. 

\subsubsection{Recommendations.} Measuring covariate complexity with annotated data is more accessible and scalable than doing user studies. Although we think that measuring the correspondence of prototypical parts with ground-truth annotations (e.g. object parts) is valuable, it is important to be aware of its limitations. A meaningful prototype may not correspond to an annotated concept in the dataset, but could still have a semantic meaning, such as representing a particular colour, texture or shape. Hence, these automated purity metrics relate closely to the Coherence property~(Section~\ref{sec:prop:coherence}). User studies or manual visual inspection could complement automated evaluation. For example, intruder detection experiments can be organised. The high detection accuracy in the user study by~\cite{nips/GhorbaniWZK19} confirm that their shown image segments were coherent to humans, but also indicate that an intruder detection experiment might be too straightforward. We see possibilities to make the evaluation more insightful by showing e.g. image patches with a lower but still sufficiently high similarity score for a prototype, or by showing an intruder that has a low but non-zero similarity to the prototype. Such experiments will give more insights into the perceived `decision boundaries' of humans regarding visual similarity and prototype complexity. 

\subsection{Compactness}
\label{sec:prop:compactness}
{Compactness describes the size of the explanation since an explanation should not overwhelm the user.
For prototypical part models, compactness is usually evaluated by counting the number of prototypes. E.g. ProtoTree~\cite{Nauta_2021_CVPR_ProtoTree} and ProtoPool~\cite{rymarczyk_2022_protopool} have roughly $10\times$ fewer prototypes than ProtoPNet. It is important to distinguish between the number of prototypes in a \emph{local} and \emph{global} explanation. The global explanation shows the full classification model, while the local explanation shows the model's reasoning for a particular input image. A local explanation of a part-prototype model is thus always a subset of the global explanation. E.g., a local explanation in ProtoTree is a particular path in the tree and hence substantially reduces explanation size compared to showing the whole tree. The most suitable explanation size can depend on the user and the intended task. E.g., Kim et al.~\cite{Kim2022HIVE} only showed the last two prototypes in a decision path of ProtoTree to its participants in a user study, to reduce their cognitive workload. In addition to measure the size of the explanation, also \emph{redundancy} of the explanation can be evaluated~\cite{Nauta2023_csur_evaluating-xai-survey}. For part-prototype models, it is desired that prototypes complement each other and are not redundant.

\subsubsection{Related Work.} 
Sinhamahapatra et al.~\cite{sinhamahapatra2022towards} manually evaluate the redundancy between prototypes and they argue that prototypes should be ``semantically disentangled'', meaning that ``each prototype should represent distinctly different semantic units''. They evaluate this disentanglement with a manual visual inspection of the prototypes, and find that ProtoPNet contains many redundant prototypes, whereas ProtoTree avoids redundancy due to the substantially lower number of prototypes, resulting in prototypes that are ``quite semantically disentangled over the whole dataset''~\cite{sinhamahapatra2022towards}. This is in line with other findings, which quantified that over 200 prototypes in ProtoPNet are identical or visually similar~\cite{nauta_exproto_2021}. ProtoPFormer~\cite{xue2022protopformer} and PIP-Net~\cite{nauta_pipnet} explicitly optimise for the diversity between prototypical parts in order to prevent redundancy. 

\subsubsection{Recommendations.} 
Compact explanations are desired, although the optimal size may depend on the number of discriminative features needed for the classification task, and the cognitive load and time availability of the user. However, evaluating compactness alone would be insufficient for quantifying explanation quality, since also the interpretability of the prototypes should be taken into account. Consider the simple example where a classifier should detect whether there is a `sun or dog' present in the image~\cite{nauta_pipnet}. The resulting classifier could theoretically contain a single prototype that encodes both the sun \emph{and} the dog. Such a model would score high in terms of compactness, but the prototypes do not encode clear, unambiguous semantically meaningful concepts and would therefore conflict with the Coherence property. Rather, a model with two prototypes (one for sun and one for dog) would be easier to understand. This example hence motivates again that explainability is a multi-faceted property that should be evaluated from multiple dimensions. 

\subsection{Composition}
\label{sec:prop:composition}
Composition evaluates the presentation format and organisation of the explanation, and focuses on \emph{how} something is explained rather than \emph{what} is explained. 
Composition can be evaluated by comparing different explanation formats with the same content, or by asking users about their preferences regarding the presentation and structure of the explanation.

\subsubsection{Related Work.} 
The comprehensibility of decision tables, trees and rule-based models was compared by~\cite{huysmans_empirical_2011}, and Jeyakumar et al.~\cite{nips/JeyakumarNCGS20} did a large-scale user study to determine which explanation style users prefer for understanding DNN model decisions. Although they did not evaluate \emph{part}-prototype models, they compared heatmaps and feature importance methods with an explanation method that presents full data samples from the dataset as representative samples~\cite{nips/JeyakumarNCGS20}. They found that this explanation-by-example style was most preferred for image classification, which is a promising result for part-prototype models. Also Kim et al.~\cite{Kim2022HIVE} evaluated explanation forms and compared part-prototypes from ProtoTree~\cite{Nauta_2021_CVPR_ProtoTree} and ProtoPNet~\cite{nips/ChenLTBRS19} with heatmaps from GradCAM~\cite{iccv/SelvarajuCDVPB17} and BagNet~\cite{bagnet_iclr}. They found that participants self-rated their level of understanding similarly, between 3 (fair) and 4 (good), for both explanation formats.  

\subsubsection{Recommendations.} 
In addition to comparing explanation styles, future research could study how part-prototypes can be best presented to the user. For ProtoPNet, it was reported that a `standalone' prototype, visualised as a single image patch, can be difficult to interpret~\cite{sinhamahapatra2022towards}. Instead, PIP-Net~\cite{nauta_pipnet} visualises a prototype as a set of multiple image patches. In  addition to assessing the most suitable composition for visualising a part-prototype, a future study could analyse how these prototypes can be best structured and included in the reasoning process. For part-prototype models, we can distinguish between a fully-connected linear layer of prototypes as used in ProtoPNet, a sparse linear layer in PIP-Net and a decision tree with prototypes as in ProtoTree. A future research question would be which format a user prefers when the part-prototypes are the same. 

\subsection{Confidence}
\label{sec:prop:confidence}
Confidence describes whether there is probability information in the explanation regarding the confidence or uncertainty of the explanation or model output. 
The confidence property is mostly a tick mark indicating whether the predictive model and/or the explanation provide confidence estimates. 
The accuracy and reliability of these confidence/uncertainty estimates can be evaluated by comparing with random confidence estimates, and by calculating the correlation with feature deletion metrics~\cite{Nauta2023_csur_evaluating-xai-survey}. 

\subsubsection{Related Work.} 
Most part-prototype models are trained with softmax and cross-entropy loss, such that the model's classification confidence can be reported. E.g., ProtoPNet~\cite{nips/ChenLTBRS19} and ProtoPShare~\cite{protopshare_rymarczyk_2021} use softmax to normalise raw output logits to probabilities, and ProtoTree~\cite{Nauta_2021_CVPR_ProtoTree} applies softmax for the class distributions in the leaves. Instead, PIP-Net~\cite{nauta_pipnet} does not provide classification probabilities at inference, but rather generates unnormalised scores. 

\subsubsection{Recommendations.} 
We recommend to distinguish between \emph{classification} confidence, \emph{explanation} confidence, and \emph{out-of-distribution detection} confidence, which can all be evaluated in future work. Softmax is known to provide over-confident class probabilities~\cite{guo_calibration_2017}, motivating a future study that analyses the reliability of a model's classification probabilities. Additionally, it would be relevant to extract \emph{explanation} confidence that indicates how confident the explanation generation method is, e.g. of the similarity calculation or visualisation process. Prototype similarity scores may reveal information about the model's uncertainty, and we are in favour of further research that would investigate whether explicit explanation confidence values can be generated \emph{in addition} to classification confidence. A user would then get a deeper understanding of the model's uncertainty, and can take over automated decisions when the model is not confident enough. 

Additionally, softmax probabilities can fail to decrease for input data that is far from the training distribution~\cite{pearce2021understanding}. We therefore expect that most part-prototype models will be overconfident for out-of-distribution (OoD) data, which should be assessed by a future study. Instead, PIP-Net~\cite{nauta_pipnet} does not provide classification probabilities and is specifically designed to handle OoD data by using a scoring-sheet reasoning and outputting near-zero scores for OoD images. Other models with scoring-sheet reasoning, such as ProtoPNet and ProtoPShare~\cite{protopshare_rymarczyk_2021}, normalise output logits with softmax and will therefore not be able to detect OoD data and accurately generate confidence estimates. However, we hypothesise that the scale of the raw output scores might be an indication of the model's classification confidence. Tree-shaped reasoning such as ProtoTree might be more problematic, since an OoD image provided to ProtoTree will be classified according to the most-left leaf when no prototype was found in the image. A `quick fix' might be to make the most-left leaf untrainable such that the model abstains from classifying images in that leaf. However, we think more suitable approaches to let part-prototype trees deal with OoD data and provide precise uncertainty estimates would be possible. Concluding, further research should investigate the reliability of different confidence outputs, and analyse to what extent part-prototype models can handle OoD data. We are convinced that reliable uncertainty estimates will contribute to more intuitive and trustworthy explanations. 

\subsection{Context}
\label{sec:prop:context}
Context addresses the extent to which the user and their needs are taken into account for comprehensible explanations that are relevant to the user’s needs and level of expertise.
Most evaluation methods are quantitative and do not consider the end-user in their relevant context~\cite{Nauta2023_csur_evaluating-xai-survey}. Quoting Colin et al.~\cite{colin_what_2022}: ``it is not yet clear (1) how useful current explainability methods are in real-world scenarios; and (2) whether current performance metrics accurately reflect the usefulness of explanation methods for the end user''. Evaluation approaches addressing the Context property are therefore highly relevant for ``application-grounded''~\cite{doshi-velez_considerations_2018} evaluation, especially since different types of users may have different perceptions and explanatory values~\cite{ehsan2021explainable}. 

\subsubsection{Related Work.} 
Colin et al.~\cite{colin_what_2022} conducted psychophysics experiments to evaluate the usefulness of heatmaps in multiple real-world scenarios and found that heatmaps can help end-users in detecting biases and identifying a model's classification strategy but are not helpful for understanding failures.
In a user study with 15 participants on usefulness of prototypes learned by ProtoTree and ProtoPNet, it was found that only 20\% of the prototypes of ProtoTree and 27\% of the prototypes of ProtoPNet were found totally useful for identifying the class~\cite{sinhamahapatra2022towards}.
Kim et al~\cite{Kim2022HIVE} performed a more objective user study (50 participants per experiment) to evaluate the usefulness and practical effectiveness of visual explanations, including ProtoTree and ProtoPNet, in AI-assisted decision making. Concretely, they investigated how useful explanations are for detecting whether a model's prediction is correct, and found that participants score above the random baseline but are far from the perfect score, indicating room for improvement. Participants were however lay crowdworkers, and evaluating with domain experts or end-users in a real-world application was left for future work. Nguyen et al.~\cite{nguyen_effectiveness_2021} did a user study to evaluate whether explanations can successfully assist users with image classification and model debugging, and compared heatmaps with showing nearest training-set examples. They found that AI-experts performed substantially better with nearest neighbour explanations than heatmaps, which is a promising result for part-prototype image patches. 

\subsubsection{Recommendations.} 
We identify research opportunities to evaluate part-proto\-type models with application-grounded user studies, similar to evaluation with heatmaps~\cite{colin_what_2022}. These evaluations can be subjective or objectively measured. An often applied evaluation method is to ask users to self-rate explanations on properties as usefulness, trust and relevance~\cite{Nauta2023_csur_evaluating-xai-survey}. However, cognitive biases and response biases often lead to discrepancies between self-reported results and findings from behavioural experiments~\cite{williams2017measuring}, and this has also been observed in the context of XAI. E.g., Hase and Bansal~\cite{acl/HaseB20} collected subjective ratings for the quality of explanations from a prototype model for text and tabular data, and found that participant's ratings were not predictive for their performance on forward or counterfactual simulatability tasks. 
Moreover, Kim et al.~\cite{Kim2022HIVE} warn for a \emph{confirmation bias} where participants prefer a model with explanations over a model without explanations, and tend to believe that a model's prediction is correct when explanations are provided. Also \cite{kaur_interpreting_2020} found that the existence of explanation can already lead to over-trust, although user's trust after showing explanations may depend on the accuracy of the predictive model~\cite{papenmeier_complicated_2022}. Results from a user study should therefore be carefully interpreted, and ideally these risks are already taken into account when setting up the study design. 

\subsection{Coherence}
\label{sec:prop:coherence}
Coherence describes how coherent the explanation is with prior knowledge and beliefs, and addresses the plausibility or reasonableness of explanations to users.
Coherence is often evaluated with anecdotal evidence by visualising example prototypes, as done by e.g. ProtoTree~\cite{Nauta_2021_CVPR_ProtoTree}, PIP-Net~\cite{nauta_pipnet}, ProtoPNet~\cite{nips/ChenLTBRS19} and ProtoPool~\cite{rymarczyk_2022_protopool}. For a more objective evaluation, annotated data can be used to automatically evaluate how well prototypes align with domain knowledge. 

\subsubsection{Related Work.} 
Goyal et al.~\cite{goyal2019explaining} calculated how often discriminative regions in counterfactual visual explanations lie near object part keypoints, and concluded that the selected counterfactual image patches are often ``semantically meaningful''. Similarly, the purity of prototypes in PIP-Net~\cite{nauta_pipnet} is evaluated by calculating the overlap between prototypical image patches and object part annotations. 
Xu-Darme et al.~\cite{xudarme_romain_sanity_2023} did not evaluate with prototypical \emph{parts} but rather measured the intersection of prototypical image patches with the segmentation of the main object. They find that the original visualisation method of ProtoPNet and ProtoTree (bicubic upsampling) results in 35\% of the prototypes in ProtoTree not having overlap with a bird object segmentation mask\footnote{In ProtoPNet, only 2\% of the prototypes have no overlap with an object, since ProtoPNet uses cropped images which makes it less likely to entirely miss the object.}. However, when applying the new PRP visualisation method~\cite{gautam_prp_2023}, only 0.5\% of ProtoTree's prototypes do not have any overlap, outperforming ProtoPNet. Rather than only evaluating how well prototypes align with a certain object, prototypes can be 
constrained to concentrate on foreground patches~\cite{xue2022protopformer}, to optimise for Coherence and Covariate Complexity directly. 
However, these automated evaluation methods for calculating the correspondence with data annotations have been shown to correlate poorly with participant's behaviour in user studies~\cite{nguyen_effectiveness_2021,Kim2022HIVE,colin_what_2022}. For example, Kim et al.~\cite{Kim2022HIVE} found near zero correlation between the localisation quality of heatmaps and participant's confidence in forward simulatability experiments. These discrepancies again underline the importance of evaluating explanation quality from multiple dimensions and complementing automated evaluation with application-grounded evaluation with user studies. 

ProtoTree and ProtoPNet were also applied on a synthetic dataset to \emph{manually} evaluate whether the model's reasoning was similar to human logic~\cite{sinhamahapatra2022towards}. One of their findings is that ProtoTree classifies a class based on \emph{absence} of prototypes, which can be different from a human's classification process. Rymarczyk et al.~\cite{rymarczyk_2022_protopool} asked participants how certain they were that a prototypical image patch was discriminative for classifying a given object. We emphasise that such questions do not evaluate the correctness of the model's reasoning, but only quantify whether the shown image patches are in line with expectations and prior knowledge of the participants. They reported that users found prototypes from their ProtoPool method (including their ``focal similarity'') often more distinctive than prototypes from ProtoTree~\cite{rymarczyk_2022_protopool}. 

\subsubsection{Recommendations.}
Prototypes are often evaluated based on anecdotal evidence, with automated evaluation with an annotated dataset, or with manual evaluation. Other quantitative evaluation methods with user studies would be to measure \emph{subjective satisfaction} and \emph{subjective comparison}, where participants can rate their satisfaction, preference and trust for part-prototypes~\cite{Nauta2023_csur_evaluating-xai-survey}. Future work could therefore more extensively capture the subjective dimension of XAI evaluation to address the Coherence property, supported by e.g. already developed XAI questionnaires~\cite{hoffman_metrics_2019}. The resulting insights could in turn provide suggestions to improve part-prototype models further.

\subsection{Controllability}
\label{sec:prop:controllability}
Controllability describes how interactive/controllable an explanation is to a user. 
Existing part-prototype models are in principle static models that are user-independent. Of course a designer, or users themselves, can choose which prototypes are shown (e.g. a global or local explanation, or only the top-k relevant prototypes). For an improved user experience, a graphical user interface can be added for visualising the reasoning of a part-prototype model. Such an interface could also enable personalised explanations where users can interact with the model and ask counterfactual questions. A promising direction that exploits the interpretability of part-prototype models, is that the user would have the possibility to directly \emph{manipulate} the explanation and, in turn, the model's reasoning.  

\subsubsection{Related Work.} 
Kulesza et al.~\cite{kulesza_principles_2015} introduced \emph{explanatory debugging}, an approach where the user can personalise a prediction model by means of its explanations. Allowing a user to disable undesired prototypes would be a concrete and promising example of explanatory debugging for part-prototype models.

\subsubsection{Recommendations.} 
Since part-prototype models are interpretable by design, adapting the explanation implies that also the model's reasoning is adapted. We identify research opportunities to enable users to suppress or modify learned prototypes, ideally supported by a graphical user interface. Such a human-in-the-loop approach may allow `fixing' the model by removing spurious correlations and aligning it more with human reasoning. The \emph{human feedback impact}~\cite{Nauta2023_csur_evaluating-xai-survey} evaluation method can then measure whether the model accuracy and explanation quality improve after human involvement. Being able to directly control and manipulate a model's reasoning is a strong benefit of part-prototype models, and is therefore a promising direction that should be investigated further.

\section{Discussion and Broader Impact}
\label{sec:discussion}

\paragraph{Importance of part-prototype models}
Explainable and interpretable models are important for responsible AI, especially in high-risk decision domains~\cite{bruckert2020next}. Explainable AI could reveal that an accurate prediction model is right for the \emph{wrong} reasons due to biases or shortcuts~\cite{rudin2019stop,Nauta2023_csur_evaluating-xai-survey}. Where most explanations only give an approximation of the reasoning, part-prototype models are interpretable by design and give insight into their full decision making process~\cite{nauta_pipnet}. More and more use cases for interpretable-by-design models in the medical domain are emerging, e.g. part-prototype models are used for breast cancer detection in mammograms~\cite{carloni_2022,wang2022knowledge}, COVID-19 detection in CT-scans~\cite{Singh2022-proto-on-chest_ctscan} and Alzheimer detection in MRI scans~\cite{mohammadjafari2021using}. Explanations for automated decisions can also become a legal requirement, such as the European GDPR law demanding ``meaningful information about the logic involved''~\cite{goodman2017european}, and the upcoming EU AI act. 

Due to the built-in interpretability of part-prototype models, the relatively new research area of part-prototype models holds promise for many applications. The literature on part-prototype models is however fragmented. In this work, we collect and organise findings from existing work that evaluate particular aspects of part-prototype models using a bottom-up analysis. This work therefore acts as a roadmap for researchers seeking to explore existing part-prototype literature.

\paragraph{Evaluation paradigm} Evaluation of explainable AI methods is important to ensure that explanations are truthful, relevant and understandable. As opposed to other communities, such as information retrieval and machine learning, the XAI community does not have an agreement on evaluating those methods. Recently, multiple suggestions for evaluating XAI have been proposed~\cite{Nauta2023_csur_evaluating-xai-survey,vilone2021notions,zhou_evaluating_2021,clement2023,hoffman_metrics_2019}. XAI evaluation papers mostly address post-hoc XAI methods, that approximate an already trained prediction model. Since part-prototype models are interpretable by design, existing evaluation methods for post-model XAI may not be directly applicable to in-model explainability methods. We fill a gap in the XAI evaluation landscape by providing a detailed analysis of XAI evaluation for interpretable part-prototype models.

Our contribution on part-prototype model evaluation is three-fold. First, we assess evaluation methods in XAI more broadly and discuss their applicability to part-prototype models. Our structured approach along the Co-12 properties puts existing works in a framework, shows that part-prototype models can be evaluated from various dimensions, and helps to identify research gaps. Second, we take the specific nature of part-prototype models into account by outlining how existing evaluation methods should be adapted to make them compatible and relevant for part-prototype models, and also suggest new evaluation methods for part-prototype models such as the target-sensitivity of the location of part-prototypes. Thorough evaluation of part-prototype models can reveal their strengths and limitations, and in turn accelerate progress and innovation. Third, we contribute to consistency, awareness and clarity in terminology for XAI evaluation. For example, we distinguish between three types of confidence, and discuss both model-output-completeness and human-output-completeness. 

In this way, we facilitate effective communication, efficiently guide future work for a more comprehensive evaluation and contribute to a solid foundation for further advancements in the field. Our evaluation cheat sheet (Table~\ref{tab:co12_cheat_sheet_proto}) is a concise overview of evaluation methods for part-prototype models, which can serve as a guideline for method developers and as a discussion point for the community for future improvements. Our suggested evaluation approaches include both content-related, presentation-related and user-related experiments, and are therefore relevant for various disciplines, including computer science, human-computer interaction (HCI) and user experience design.

\section{Conclusion}
\label{sec:conclusion}
We have discussed recent work that has evaluated various Co-12 properties for part-prototype models, and have also made concrete recommendations for further evaluation, as summarised in our Co-12 cheat sheet in Table~\ref{tab:co12_cheat_sheet_proto}. We agree with Jacovi \& Goldberg~\cite{acl/JacoviG20} that interpretable methods should be held to the same standards as post-hoc explanation methods. With the Co-12 recipe presented in this paper, we aim to contribute to the progression and maturity of the relatively new research field on interpretable part-prototype learning. 

\clearpage
\bibliographystyle{abbrv}
\bibliography{references}

\end{document}